\documentclass[letterpaper, 10 pt, conference]{ieeeconf}  
\IEEEoverridecommandlockouts                              
\usepackage{amsmath} 
\usepackage{xcolor}
\usepackage{booktabs}
\usepackage{graphicx}
\usepackage{capt-of}
\usepackage{microtype}
\usepackage{url}

\title{\LARGE \bf
SAFE: Unified Slip and Fracture Detection with Low-Cost Acoustic Sensing in Robotic Grasping
}
\author{Zerun Wang$^{1, 2}$, Vivek Kamat$^{2, *}$, and Shekhar Bhansali$^{2}$
\thanks{$^{*}$Corresponding author.}%
\thanks{$^{1}$School of Electrical and Computer Engineering, Georgia Institute of Technology, Atlanta, GA 30332, USA. {\tt\footnotesize zerunwang@gatech.edu}}%
\thanks{$^{2}$Department of Electrical and Computer Engineering, Vanderbilt University, Nashville, TN 37240, USA. \newline
        {\tt\footnotesize \{vivek.a.kamat,shekhar.bhansali\}@vanderbilt.edu}}%
}

\makeatletter
\let\@oldmaketitle\@maketitle
\renewcommand{\@maketitle}{\@oldmaketitle 
  \begin{center}
    \includegraphics[width=0.97\textwidth]{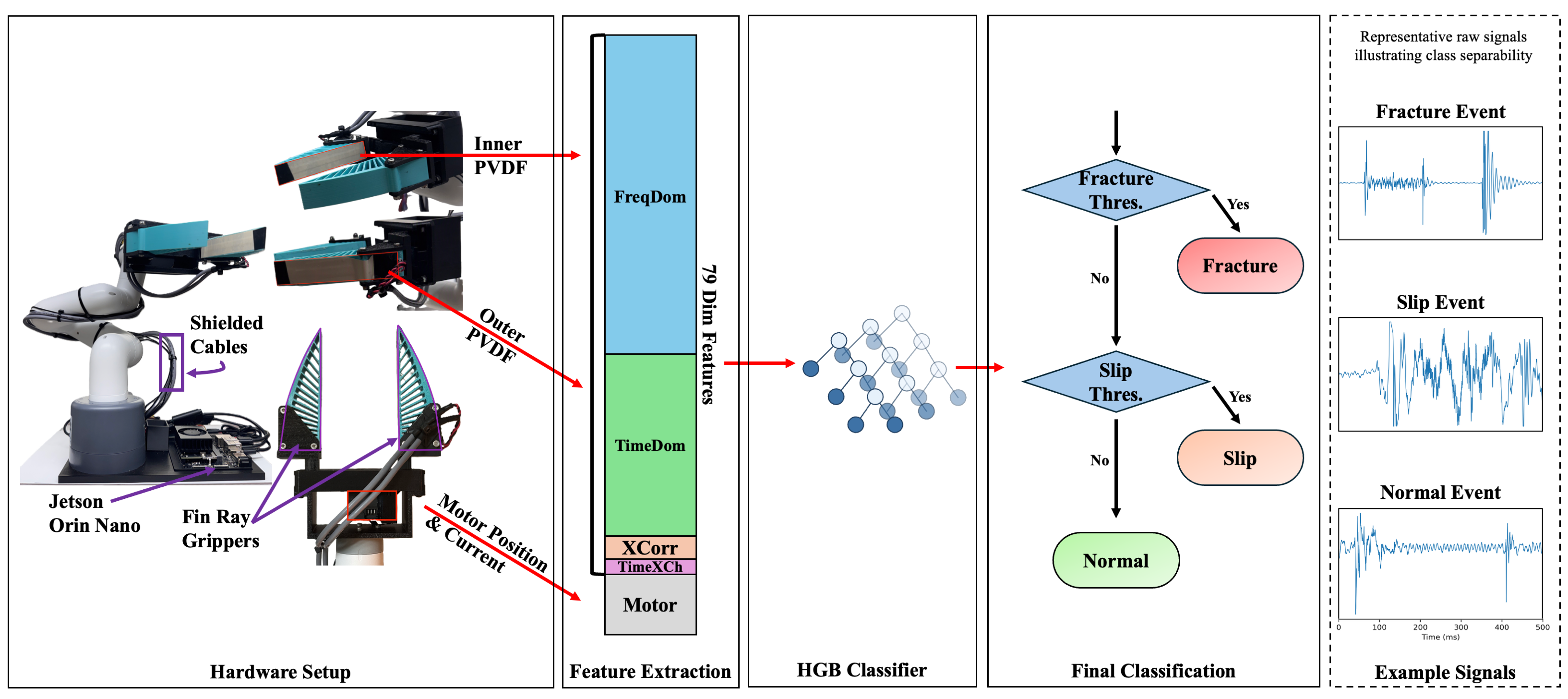}
    \vspace{-0.4cm}
    \captionof{figure}{System overview. Acoustic signals from two passive PVDF sensors mounted on the inner and outer surfaces of a compliant Fin Ray finger, together with motor current and position data, are acquired on a single NVIDIA Jetson Orin Nano. Each 500-ms window is converted into a 79-dimensional feature vector and passed to a HistGradientBoosting (HGB) classifier. A fixed evaluation cascade guarantees a single deterministic state (normal, slip, or fracture) per window. Rightmost panel: representative raw signals for each class, illustrating their distinct acoustic signatures.}
    
    \label{fig:system_overview}
  \end{center}
  \vspace{-0.4cm}
}
\makeatother

\begin{document}

\maketitle
\addtocounter{figure}{-1}
\thispagestyle{empty}
\pagestyle{empty}

\begin{abstract}
Manipulating fragile objects remains challenging as robots must understand the state of what they grasp, such as slip or fracture, to respond appropriately, especially when material properties are unknown. In this paper, we present SAFE: a low-cost, general-purpose sensing approach that detects both slip and fracture in real time using two passive polyvinylidene fluoride (PVDF) acoustic sensors and motor proprioception, without relying on vision or prior material knowledge. The sensors are mounted on a compliant Fin Ray gripper, and a unified HistGradientBoosting classifier reports the state (normal, slip, or fracture) from a 79-dimensional feature vector. Under leave-one-grasp-out cross-validation, SAFE achieves an Alert-F1 of 0.884 with near-zero slip-fracture confusion, and ablations confirm that acoustic sensing is indispensable. An adaptive grasp controller built on this detection layer runs at 104~Hz on a Jetson Orin Nano, achieving 91.3\% success across 46 closed-loop robot trials spanning diverse object categories, while each fixed-force strategy drops to 0\% on object conditions that mismatch its preset. It further reaches 82.4\% success on novel objects unseen during training, demonstrating robust, failure-aware grasp control without object-specific calibration.
\end{abstract}

\section{Introduction}
Robotic manipulation of fragile objects requires a robot to recognize two very different kinds of failure: the object slipping out of the gripper, and the object fracturing inside it. Slip has received considerable attention, since it can be inferred from force, vibration, or visual cues and corrected before the object is completely dropped. Fracture is a harder problem and a comparatively underexplored one. A sensing and control system that only handles slip, or only handles fracture, addresses half the problem. Fully characterizing the safe grasping range requires estimating both the lower boundary (slip) and the upper boundary (fracture). Relying on slip feedback alone creates a trade-off between speed and safety: reaching a stable grip quickly requires larger force increments, which raises the risk of fracturing the object before slip is even detected. 

Detection means something different for each failure mode. For slip, there is typically a reaction window between the onset of relative motion and the object fully leaving the gripper, giving a controller time to increase force before the grasp is lost. For fracture, detection instead marks a state in which a crack is already present. What detection provides here is not prevention of that first instant of damage, but the ability to halt further force escalation and contain it before it propagates, rather than continuing to apply an escalating grasp force under the false assumption that the object remains intact. This distinction matters more as robots move into complicated environments, such as household settings, where the same robot needs to execute a sequence of real-time-decided manipulations. Relying purely on pre-tuned, fixed-force policies becomes increasingly limiting in this setting. If the first step in a sequence induces slip, the next step must adjust its grasp accordingly, whereas if the first step fractures the object, whether to continue the original plan at all becomes a decision the robot must be able to make, rather than one it can silently ignore.

Existing approaches to this problem generally fall into two lines of work, which we review in detail in Section~\ref{sec:related_works}. Directly instrumenting the gripper, whether with multimodal sensors~\cite{waltersson2026deformable}, reactive tactile force control~\cite{ayral2026reactive}, or pneumatic and barometric channels~\cite{shang2026forte}, has produced effective slip-regulating controllers, but comparatively little work treats fracture as a distinct, detectable failure mode rather than an incidental force spike~\cite{nishimura2019fluid, ishikawa2022learning, yagawa2023learning}, and such hardware is often tailored to a specific gripper design. On the acoustic sensing side, high-frequency vibrations are directly emitted during fracture initiation and can be used for real-time monitoring~\cite{chuluunbat2018acoustic}. Existing PVDF-based acoustic sensing methods have mainly focused on detecting slip-induced vibrations using handcrafted signal processing and threshold-based decision rules~\cite{Nikafrooz2022, Rosle2025}. Whether the same sensing modality can also distinguish fracture events, and whether a unified learned model can reliably recognize both slip and fracture for real-time closed-loop control, remains unexplored.

We present SAFE (\textbf{S}lip \textbf{A}nd \textbf{F}racture d\textbf{E}tection), a framework that answers this question with a deliberately minimal sensing layer containing two passive PVDF acoustic films mounted on opposite surfaces of a Fin Ray finger, combined with the motor's own current and position feedback. Unlike prior acoustic slip detectors, SAFE is trained as a unified three-class classifier that distinguishes slip, fracture, and normal events from one 79-dimensional feature vector, running at approximately 104~Hz on a single NVIDIA Jetson Orin Nano. Because the sensing hardware is passive, low-cost, and adds no meaningful stiffness to the finger, and the classifier requires no additional compute or actuation hardware, SAFE itself can provide sufficient data for a closed-loop adaptive grasp controller and can also serve as an add-on providing additional grasping state data to other grasping platforms. Evaluated under leave-one-grasp-out cross-validation, SAFE achieves an Alert-F1 of 0.884 with near-zero confusion between slip and fracture, and the resulting adaptive controller reaches a 91.3\% closed-loop grasp success rate across 46 real-robot trials, compared to 71.7\% and 60.9\% for fixed-force baselines. It further reaches 82.4\% success across 51 trials on object categories entirely absent from training, showing that the same sensing layer generalizes beyond the specific objects it was trained on.

The contributions of this work are as follows:
\begin{itemize}
    \item A low-cost, two-channel passive acoustic sensing scheme, mounted on a Fin Ray finger, shown through ablation to be required for reliable slip and fracture detection;
    \item A unified three-class classifier that distinguishes slip, fracture, and normal contact with near-zero cross-confusion (3 of 437 alert windows assigned to the opposite failure mode), evaluated under a leave-one-grasp-out protocol that avoids trial-level leakage;
    \item An adaptive grasp controller that integrates this detection layer in closed loop, achieving 91.3\% grasp success on five object categories, compared to 71.7\% and 60.9\% for fixed-force baselines, and 82.4\% success across 51 trials on three object categories entirely absent from the training data.
\end{itemize}

\section{Related Works}
\label{sec:related_works}
\subsection{Tactile Sensing for Contact-Rich Manipulation}
Tactile sensing allows robots to feel forces, textures, and object shapes, providing contact information that is unavailable from vision alone, and has become a key enabling technology for contact-rich manipulation. Visual-tactile sensors reconstruct contact force and geometry from deformable membranes inside grippers at high spatial resolution~\cite{yuan2017gelsight, lambeta2020digit}, and have been integrated into Fin Ray structures through embedded cameras that capture the contact surface directly~\cite{liu2022gelsight, liang2025alltact}, as well as through mechanical transmission of structural deformation~\cite{nam2025tacfinray}. Most notably, GelSight FlexiRay achieves full-coverage multimodal perception, including proprioception, contact force, texture, and slippage, on Fin Ray grippers using a multi-mirror visual-tactile design~\cite{wang2024gelsight}. However, such vision-based tactile approaches face optical and imaging trade-offs: temporal responsiveness is bounded by the frame rate of the imaging sensor, while structural form factor is constrained by the camera's field of view and focal length~\cite{wang2024gelsight, li2024when}. Barometric sensing offers an alternative: integrated into Fin Ray grippers, it enables force and slip estimation with low latency, but pneumatic channels remain vulnerable to temperature-induced pressure drift, limiting deployment in thermally variable scenarios~\cite{shang2026forte}.

While comprehensive tactile reconstruction benefits general-purpose manipulation, many grasp-control tasks do not require full contact-state estimation. Instead, reliably recognizing a small number of task-critical interaction events can be sufficient for closed-loop grasp control. For fragile object manipulation with unknown material properties, two such events define the safe grasping window from opposite directions: fracture, when grasp force exceeds what the object can withstand, and slip, when grasp force falls short of what is needed to hold it. 

\subsection{Fracture Detection}
Excessive grasp force can cause irreversible damage, making timely fracture detection critical for safe grasp-force regulation. Early approaches monitor contact pressure using compliant fluid fingertips to identify pressure changes and regulate grasp force to avoid damage~\cite{nishimura2019fluid}. More recent learning-based methods exploit tactile observations to anticipate impending fracture during robotic food manipulation~\cite{ishikawa2022learning}, with follow-up work improving object generalization by learning fracture-related representations from reusable surrogate objects rather than task-specific fragile items~\cite{yagawa2023learning}. These studies demonstrate the value of tactile sensing for reducing grasp-induced damage, but they also expose two limitations. First, existing approaches rely primarily on pressure or deformation sensing, whereas acoustic-emission studies have shown that crack initiation and propagation generate high-frequency elastic waves, with AE activity emerging before yielding and characteristic burst signals accompanying fracture initiation and crack growth~\cite{chuluunbat2018acoustic,melchiorre2023acoustic}. Such signals provide sensitive indicators of damage evolution before macroscopic failure becomes apparent. Second, many tactile sensing approaches rely on relatively rigid taxel arrays, which can complicate integration with highly compliant end-effectors such as Fin Ray grippers.

More fundamentally, all of the above methods address only one half of the grasp-force problem: they provide an upper bound that prevents excessive force, but say nothing about whether the applied force is sufficient to hold the object. Successful grasping of unfamiliar fragile objects requires both avoiding fracture and avoiding slip, and fracture-only perception cannot by itself determine an appropriate grasp force.

\subsection{Slip Detection}
Symmetric to the fracture case, insufficient grasp force leads to object slip and grasp failure, making slip perception the complementary lower bound for adaptive grasp control. Early work demonstrates that grasp force can be regulated according to incipient slip, mimicking human grasp stabilization~\cite{romano2011humaninspired}. Subsequent studies develop tactile slip detection coupled with reactive grasp controllers based on incipient-slip estimation~\cite{dong2018maintaining, james2021slip}, and more recent learning-based methods further improve robustness and object generalization across diverse manipulation scenarios~\cite{zhao2025universal}. Most relevant to our approach, A-SLIP extends passive acoustic sensing from binary slip detection to continuous estimation of slip direction and magnitude using multi-channel piezoelectric microphones and learning-based fusion, demonstrating in closed-loop tasks that continuous slip vectors enable more reliable motion tracking and stopping behavior than binary detection alone~\cite{yoo2026aslip}. While slip direction may not be significantly useful for parallel-type grippers because they have only one closing direction, magnitude information, in principle, could support force controllers to scale grip-force increments based on the severity of observed slip for faster adjustments. However, without fracture detection limiting the upper bound, there is a risk that the increment is too large and directly exceeds the fracture limit.

\subsection{Acoustic Sensing}
The two sections above demonstrate that acoustic signals carry valuable information for grasping state sensing. Acoustic emission literature identifies high-frequency vibration as the earliest indicator of fracture, while A-SLIP and other works show that acoustic signals also encode fine-grained slip dynamics. This suggests that a single acoustic sensing layer may be capable of resolving both failure modes, a possibility that, to our knowledge, has not been tested.

Acoustic sensing captures mechanical vibrations generated during physical interaction and has been explored for a variety of robotic manipulation tasks. Passive contact microphones capture high-frequency vibrations from the contact interface, enabling continuous slip vector estimation~\cite{yoo2026aslip}, material classification~\cite{oh2025miccheck}, and contact-rich manipulation tasks~\cite{li2022see,xu2025multimodal}. In contrast, active acoustic methods emit a probing signal and infer contact state from the received acoustic response~\cite{zhang2025vibecheck}. Among acoustic transducers, polyvinylidene fluoride (PVDF) film has been widely adopted for contact acoustic acquisition in grasping, owing to its flexibility, broad frequency response, low acoustic impedance, and high sensitivity to dynamic mechanical stimuli~\cite{Xin2016, Qi2023, Zhao2026}. Unlike resistive or capacitive tactile sensors, PVDF responds to transient deformation and vibration rather than static contact force, making it well suited to dynamic tactile events such as contact transitions, impacts, and slip~\cite{Xin2016, Qi2023}. Existing studies have integrated PVDF sensors into cable-driven robotic hands for slip-triggered grasp-force regulation~\cite{Nikafrooz2022} and into compliant robotic grippers for vibration-based slip estimation~\cite{Rosle2025}.

However, every acoustic-sensing study above targets a single failure mode with a single sensing channel: slip detection via one microphone and a hand-tuned threshold, or fracture-relevant acoustic emission studied only outside robotic manipulation. None uses acoustic sensing to resolve \emph{both} slip and fracture from a shared, low-cost sensing layer, and none demonstrates that a single learned classifier can distinguish the two failure signatures reliably enough for real-time closed-loop control. This is the gap SAFE addresses: two passive PVDF channels on a single compliant Fin Ray finger, combined with motor proprioception, feeding a unified classifier that resolves both the upper (fracture) and lower (slip) bounds of the safe grasping window from one sensing modality.

\section{Methodology}
Real-time detection of slip and fracture during grasping must satisfy three constraints: it cannot rely on prior knowledge of object material, it must run within a real-time control loop on low-cost edge hardware, and it must resolve two physically different failure signatures from a single, minimal sensing layer. We address these with compliant hardware carrying acoustic/proprioceptive sensing, a windowed feature representation, and a unified classifier.

\subsection{Hardware Design}
Both slip and fracture produce transient acoustic signals that can be sensed through piezoelectric polyvinylidene fluoride (PVDF) sensors owing to their high sensitivity to mechanical stimuli. Prior frequency-domain characterization has shown that slip events alter the spectral relationships between the inner and outer acoustic signals~\cite{wang2026slip}. In addition, motor current and position data are also collected.

The gripper is a compliant Fin Ray design adapted from the Stanford UMI Gripper~\cite{chi2024universal} with a Dynamixel XL330 servo as an actuator in current-control mode, mounted on a myCobot 280 arm and fabricated using a Stratasys F120 and Bambu Lab P1S 3D printers. The fingers are 3D-printed in TPU85A (100\% infill), providing the passive compliance ideal for delicate object manipulation. Two PVDF sensor films (TE Connectivity Measurement Specialties DT4-028K/L) are mounted on the inner and outer surfaces of only one finger to minimize stiffness impact, and each is connected to an identically configured channel of the OPA2134 charge amplifier ($R_f = 10\,\text{M}\Omega$, $C_f = 1\,\text{nF}$, high-pass cutoff $\approx 15.9\,\text{Hz}$). The two channels are AC-coupled to two identical USB audio interfaces sampling at 48 kHz. Both PVDF channels and motor proprioception are acquired as ROS~2 topics and timestamped using the same system clock, enabling multimodal feature extraction over aligned time intervals. Robot control, sensor acquisition, and model inference all run on a single NVIDIA Jetson Orin Nano (Ubuntu 22.04, ROS 2 Humble). The hardware overview is shown in the Hardware Setup section of Fig.~\ref{fig:system_overview}.

The complete sensing hardware, including two PVDF films, one dual-channel charge amplifier (OPA2134), and two USB audio interfaces, totals approximately \$84 in component cost, compared to \$355 for a DIGIT sensor~\cite{gelsight2026digit} or \$510 for a GelSight Mini system~\cite{gelsight2026mini}.

\subsection{Dataset Construction}
\begin{figure}[t]
    \centering
    \includegraphics[width=\linewidth]{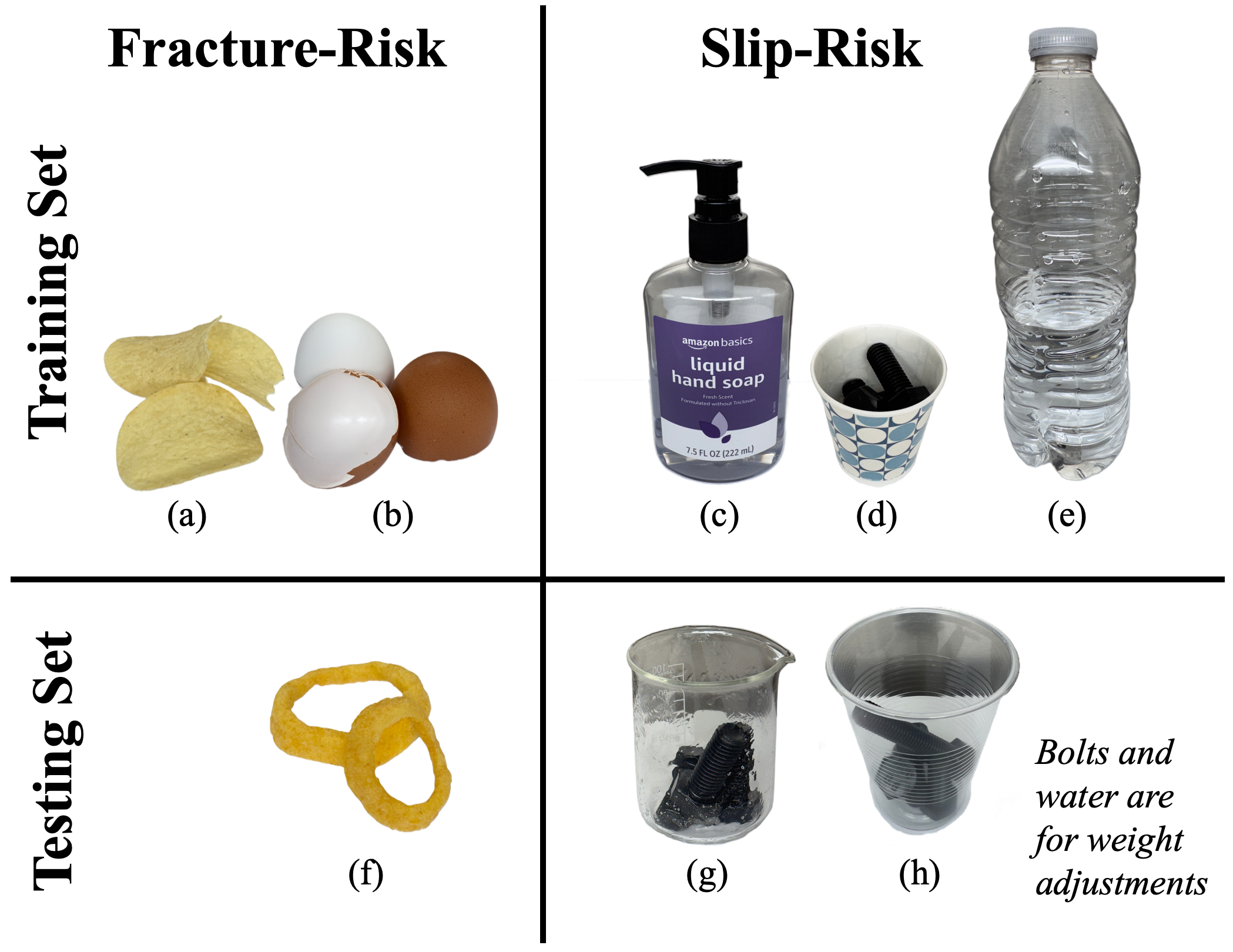}
    \vspace{-0.8cm}
    \caption{Objects used for data collection and evaluation. \textit{Training set:} two fracture-risk objects, eggshell (a) and potato chips (b), and three slip-risk objects with varied weights, hand-soap bottle (c), paper cup (d), and water bottle (e). \textit{Testing set (out-of-distribution):} onion ring chips (f), glass beaker coated with soap (g), and plastic cup (h); the latter two were collected with varied weights.}
    \label{fig:object_overview}
    \vspace{-0.5cm}
\end{figure}

Fracture events and slip events were collected separately using the procedures described below. Table~\ref{tab:dataset_composition} summarizes the resulting dataset: two brittle objects (eggshell, which fractures progressively; potato chips, which fracture instantly) for fracture data, and three objects across seven weight conditions (hand-soap bottle, paper cup, water bottle) for slip data (Fig.~\ref{fig:object_overview} upper section), together yielding 1499 labeled 500-ms windows across 252 grasp-trial recordings.

\begin{table}[b]
    \centering
    \vspace{-0.3cm}
    \caption{Training Dataset Composition}
    \label{tab:dataset_composition}
    \vspace{-0.3cm}
    \footnotesize
    \renewcommand{\arraystretch}{1.05}
    \setlength{\tabcolsep}{5pt}
    \begin{tabular}{lccc}
        \toprule
        Object & Weight (g) &
        \shortstack{Positive windows\\($n$ groups)} &
        \shortstack{Normal windows\\($n$ groups)} \\
        \midrule
        Eggshell     & --     & 216 (54) & 566 (54) \\
        Chips        & --     & 77 (35)  & 322 (39) \\
        Soap bottle  & 215.25 & 2 (1)    & 18 (9)   \\
                     & 289.84 & 62 (31)  & 20 (10)  \\
        Paper cup    & 2.29   & 4 (2)    & 36 (18)  \\
                     & 61.78  & 32 (16)  & 8 (4)    \\
                     & 68.14  & 12 (6)   & 28 (14)  \\
        Water bottle & 10.77  & 12 (6)   & 64 (32)  \\
                     & 243.51 & 20 (10)  & 0 (0)    \\
        \midrule
        \textbf{Total} & \textbf{--} & \textbf{437 (161)} & \textbf{1062 (180)} \\
        \bottomrule
    \end{tabular}
    \begin{flushleft}
    \footnotesize
    Positive windows comprise fracture and slip windows (293 from 89 fracture grasp trials and 144 from 72 slip grasp trials). A grasp-trial group may contribute both positive and normal windows. Of the 180 grasp-trial groups contributing normal windows, 89 also contribute positive windows, while the remaining 91 contribute only normal windows. Overall, the dataset contains 1499 windows from 252 unique grasp-trial groups.
    \end{flushleft}
    \vspace{-0.3cm}
\end{table}

\textbf{Fracture data collection.} The arm was held stationary throughout the whole process. Once contact with the object was detected, gripper current was gradually increased from the initial gripper closing current until the object fractured or the motor current limit was reached, with a synchronized 240~fps offline camera recording the trial for labeling purposes only. Two brittle objects were used, eggshell and potato chips. After collection, all audio segments exceeding 1.5 times the recording's median RMS energy were flagged as candidate events. By reviewing the synchronized video, each candidate event was then manually labeled as fracture or normal. This event-level labeling gave each fracture event a known start and end time, yielding 191 distinct events across 293 of the 1499 windows; events longer than 500 ms were split into multiple consecutive end-anchored windows.

\textbf{Slip data collection.} The gripper was closed under a preset fixed current and held while the arm lifted vertically approximately 15 cm within 1 second. Each trial was then manually labeled as slip or no-slip. Three object types were used, with slip induced by varying object weight and gripper current: a hand-soap bottle, a paper cup, and a water bottle. Each trial was split into exactly two fixed 500-ms windows (0–0.5 s and 0.5–1.0 s), yielding 144 slip-labeled windows. 

The remaining windows were labeled as normal: 888 from the fracture-collection recordings (gripper-closing sounds, background vibration, ordinary contact) and 174 from the slip-collection recordings, totaling 1062.

\subsection{Feature Extraction}
\begin{figure}[t]
    \vspace{0cm}
    \centering
    \includegraphics[width=0.8\linewidth]{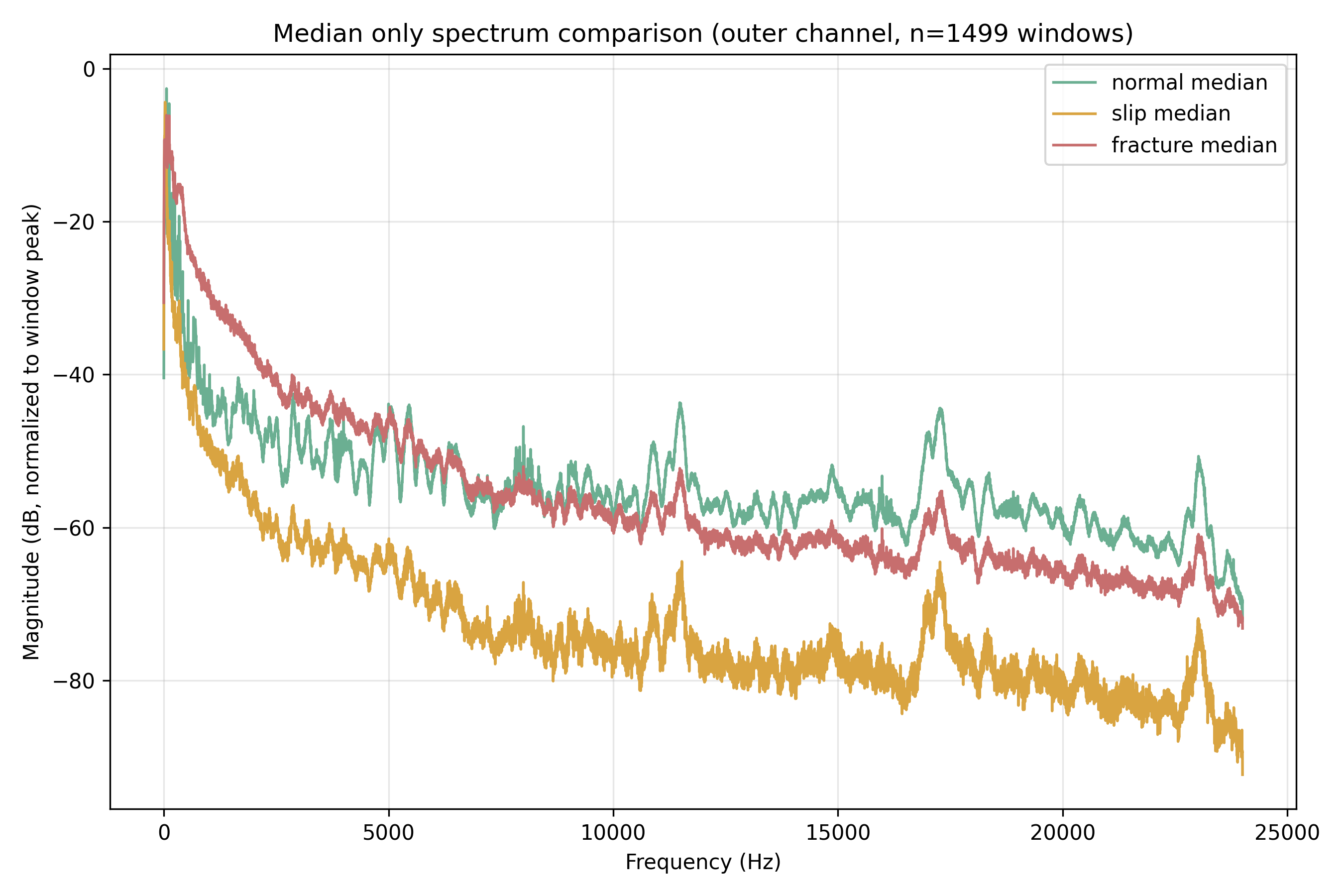}
    \vspace{-0.3cm}
    \caption{
    Median magnitude spectrum for each class on the outer channel, across 1499 windows, showing visible class separability directly in the frequency domain.
    }
    \label{fig:freq_compare}
    \vspace{-0.3cm}
\end{figure}

Fig.~\ref{fig:system_overview} (right section) and Fig.~\ref{fig:freq_compare} illustrate the time-domain and frequency-domain signals of fracture, slip, and normal events, respectively. In the time domain, fracture events exhibit sharp, discrete impulses corresponding to crack initiation and propagation, in contrast to the continuous, higher-amplitude oscillations characteristic of slip and the comparatively quiet baseline of normal contact. In the frequency domain, the median spectra of the three classes are visibly separable across most of the acquisition bandwidth, with fracture events consistently carrying more energy at low-to-mid frequencies than normal contact, and slip events consistently carrying the least. This visual separability motivates extracting both time- and frequency-domain statistics as complementary evidence for the classifier described below.

To facilitate real-time deployment and offline model training, raw sensor streams are segmented into 500-ms windows, as fracture events show a median duration of 495~ms. Specifically, at deployment, a 500-ms buffer is created for all feature extractor inputs; during offline training, each event is cropped into one or multiple windows and independently labeled with the corresponding event class.

Prior to spectral feature extraction, each window is Hann-windowed and transformed via a real-valued FFT. Each window then yields a 79-dimensional feature vector spanning five feature groups. Spectral-shape statistics (13 band-energy ratios, centroid, bandwidth, two rolloff points, dominant frequency, flatness, entropy, and a mid-to-low band energy ratio [500--2000~Hz over 0--500~Hz]; 21~dims) and time-domain statistics (RMS, peak, crest factor, log-energy, zero-crossing rate, skewness, kurtosis, attack/decay time [rise/fall time of the Hilbert envelope], envelope kurtosis, temporal centroid, and peak count; 12~dims) are each computed per PVDF channel ($42 + 24 = 66$~dims total). Cross-channel features (cross-correlation peak and lag, zero-lag correlation; 3~dims) and time-domain cross-channel ratios (peak ratio, RMS ratio; 2~dims) capture directional information. Motor current and position statistics (mean, standard deviation, range, slope, max current, and a normalized gripper-progress fraction; 8~dims) provide complementary information about applied force. In total: $66 + 3 + 2 + 8 = 79$~dimensions. 

\subsection{Model Selection and Training}
\label{sec:model_training}
HistGradientBoosting (HGB) is an ensemble of decision trees built sequentially, where each new tree is trained to correct the residual errors of previous ones. Continuous feature values are binned into histograms before each split search, which makes training and inference fast enough for real-time deployments. We use it as the proposed three-class model (fracture, slip, normal) on the full 79-dimensional feature vector. To account for class imbalance (normal comprises 71\% of samples), the classifier's built-in \texttt{class\_weight="balanced"} option is used throughout training.

Four hyperparameters (tree depth, learning rate, minimum samples per leaf, and $L_2$ regularization) were tuned via a grid search. First, 32 candidate combinations were screened using 8-fold GroupKFold cross-validation grouped by grasp trial, and the top five candidates were then re-evaluated using Leave-One-Grasp-Out (LOGO) cross-validation over the 252 grasp-trial groups. Because a single trial produces multiple 500-ms windows, grouping by trial ensures that windows recorded during the same grasp attempt are never split across the training and test sides of the same fold, which would otherwise let the model see correlated windows from one trial during training and be tested on other windows from that same trial. The five finalist configurations scored within 0.011 of each other on this evaluation, well within the model's own $\pm 0.026$ confidence interval; therefore, reusing this evaluation for both selection and reporting is unlikely to introduce substantial bias. The selected final configuration consists of a maximum depth of 5, a learning rate of 0.1, a minimum of 20 samples per leaf, and an $L_2$ regularization penalty of 0.1.

The model outputs three confidence scores jointly constrained by a single softmax. Fracture confidence is checked first against its threshold; if it does not exceed the threshold, slip confidence is checked next; both thresholds are fixed at 0.5. Given fracture and slip confidence scores $p_f, p_s \in [0,1]$ from the softmax output, the predicted state $\hat{y}$ is
\begin{equation}
\hat{y} =
\begin{cases}
\text{fracture} & \text{if } p_f \geq \tau_f \\
\text{slip} & \text{if } p_f < \tau_f \text{ and } p_s \geq \tau_s \\
\text{normal} & \text{otherwise}
\end{cases}
\end{equation}
where $\tau_f = \tau_s = 0.5$.

\subsection{Grasp Control Strategy}
\label{sec:control_policy}
The Adaptive controller is designed to converge on a grasp current that avoids both slip (too little force) and fracture (too much force). We address this with a two-phase policy.

In the first phase, the gripper closes at a small current until contact is detected (based on motor position feedback). Once contact is detected, the arm attempts a small test lift while monitoring for slip. If slip is detected during this test lift, the gripper releases slightly, returns to the pre-lift position, and recloses at 1.2 times the previous current, monitoring for fracture during the regrasp; a detected fracture aborts the trial. This test-lift-and-reclose cycle repeats until the test lift completes without slip or the current exceeds the motor limit.

In the second phase, once the object is fully lifted, releasing the grip is no longer an option as the object may drop completely, so the controller instead makes small continuous adjustments: current is decreased slightly if fracture is detected, and increased slightly if slip is detected.

\section{Experiments}
\subsection{Classification Performance}
\begin{figure}[t] 
    \vspace{0cm}
    \centering
        \begin{minipage}[c]{0.48\linewidth} 
        \centering
        \makeatletter\def\@captype{table}\makeatother
        \caption{Classification performance}
        \vspace{-0.3cm}
        \label{tab:model_performance}
        \footnotesize 
        \setlength{\tabcolsep}{3pt} 
        \begin{tabular}{lcc}
            \toprule
            Metric & Slip & Fracture \\ 
            \midrule
            Precision & 0.923 & 0.809 \\ 
            Recall & 0.910 & 0.898 \\ 
            F1 & 0.916 & 0.851 \\ 
            ROC-AUC & 0.993 & 0.968 \\ 
            PR-AUC & 0.966 & 0.851 \\ 
            \bottomrule
        \end{tabular}
    \end{minipage}
    \hfill 
    \begin{minipage}[c]{0.48\linewidth} 
        \centering
        \makeatletter\def\@captype{figure}\makeatother
        \includegraphics[width=\linewidth]{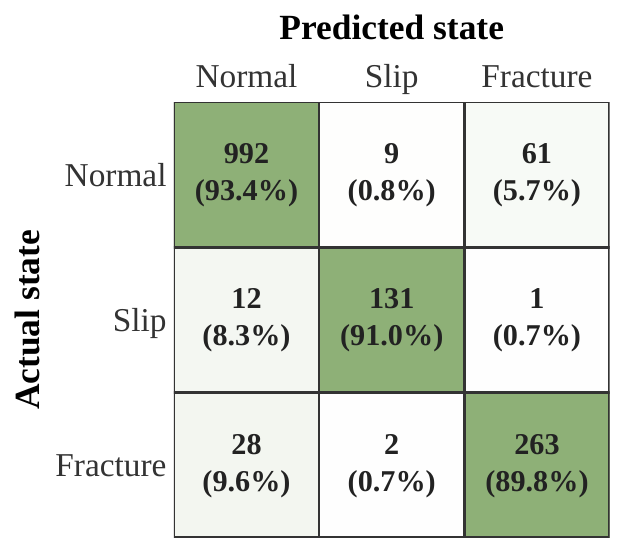}
        \vspace{-0.7cm}
        \caption{Confusion matrix} 
        \label{fig:confusion}
    \end{minipage}
\end{figure}

Table~\ref{tab:model_performance} reports classification performance under the Leave-One-Grasp-Out protocol described in Section~\ref{sec:model_training}. The proposed model achieved an Alert-F1 (the mean of slip and fracture F1: the normal class comprises 71\% of samples and would otherwise dominate an aggregate score) of 0.884 [0.856, 0.908], with slip and fracture individually reaching F1 of 0.916 and 0.851. The 95\% confidence interval (CI) was estimated using a group-level cluster bootstrap with 5000 resamples over grasp-trial groups. As shown in the confusion matrix (Fig.~\ref{fig:confusion}), the two failure modes are rarely mistaken for each other: only 3 of the 43 misclassified alert windows were assigned to the opposite failure mode, the remaining 40 being confusions with the normal class. The two failure types therefore exhibit distinguishable signatures within a shared feature space.

Across 125 deployment recordings (126,368 total inferences) on the same Jetson Orin Nano used for closed-loop control (Section~\ref{sec:closed_loop}), per-recording mean inference latency was $9.638 \pm 0.155$~ms, corresponding to a stable inference rate of approximately 104~Hz, well within the real-time budget for continuous grasp-force feedback.

\textbf{Feature contribution.} We assessed the relative contribution of each input modality both via ablation and via post-hoc importance analysis. Table~\ref{tab:ablation} reports ablation results across five feature subsets, evaluated under the identical fixed HGB configuration and 252-fold LOGO protocol used for the final model. A motor-only model performs markedly worse than any PVDF-inclusive configuration, confirming the necessity of acoustic sensing alongside the proprioceptive current and position data. Isolating the outer channel's marginal contribution (All vs. Inner PVDF + Motor) revealed a significant improvement in slip F1 ($+0.0347$, $p = 0.0164$) but no significant improvement in fracture F1 ($+0.0024$, $p = 0.8352$), where statistical significance was assessed using a paired group-level cluster bootstrap with 5000 resamples over grasp-trial groups, supporting the inclusion of both channels in the final model. Sensor-group permutation importance, evaluated under the same 252-fold LOGO protocol, corroborated the ablation results: the two PVDF channels produced substantially larger Alert-F1 drops than the motor state (inner PVDF: $0.119 \pm 0.008$; outer PVDF: $0.133 \pm 0.008$; motor: $0.048 \pm 0.008$).

\begin{table}[b]
    \centering
    \caption{Ablation study (fixed HGB config, 252-fold LOGO)}
    \label{tab:ablation}
    \vspace{-0.3cm} 
    \begin{tabular}{lcccc}
        \toprule
        Feature Set & $n$ & Alert-F1 & Slip F1 & Fracture F1 \\
        \midrule
        All (proposed) & 79 & \textbf{0.884} & \textbf{0.916} & \textbf{0.851} \\
        Inner PVDF + Motor & 41 & 0.865 & 0.881 & 0.849 \\
        Dual PVDF only & 71 & 0.829 & 0.849 & 0.809 \\
        Inner PVDF only & 33 & 0.791 & 0.788 & 0.795 \\
        Motor only & 8 & 0.667 & 0.752 & 0.582 \\
        \bottomrule
    \end{tabular}
    \vspace{-0.4cm}

\end{table}

\textbf{Architecture and baseline comparison.} We compared our proposed model with both end-to-end deep learning models and other tree architectures. As shown in Table~\ref{tab:model_comparison}, CNN and Random Forest (RF) underperformed the gradient-boosted tree model (HGB) at our dataset scale. 

We also compared against a simple decision list relying on peak acoustic features. Tuned via the same two-stage search as HGB, it performed significantly below HGB on every metric ($n=252$, $p<0.0001$). In addition, two independently trained detectors (one for slip and one for fracture) were compared with our proposed three-class model. The overall performance was statistically indistinguishable, while the binary detectors outperformed the unified model in terms of higher slip F1. We nonetheless preferred the unified model because running two separate detectors also requires two inferences per window instead of one. 

\begin{table}[t]
    \vspace{0cm}
    \centering
    \caption{Model Comparison}
    \label{tab:model_comparison}
    \vspace{-0.3cm} 
    \footnotesize 
    \setlength{\tabcolsep}{3.5pt}
    \begin{tabular}{lcccc}
        \toprule
        Model               & Alert-F1       & Slip F1        & Fracture F1    & Fracture Precision \\
        \midrule
        CNN                 & 0.614          & 0.510          & 0.718          & 0.732                    \\
        RF                  & 0.863          & 0.899          & 0.822          & 0.751                    \\
        HGB (proposed)      & \textbf{0.884}          & 0.916          & \textbf{0.851} & \textbf{0.809}           \\
        Threshold cascade   & 0.334          & 0.160          & 0.508          & 0.478                    \\
        Binary detectors    & 0.882          & \textbf{0.938}          & 0.826          & 0.783                    \\
        \bottomrule
    \end{tabular}
    \vspace{-0.5cm}
\end{table}

\subsection{Closed-Loop Evaluation}
\label{sec:closed_loop}
We compared three grasp-force controllers, each across 46 trials spanning five objects and multiple weight conditions: Fixed Low (50 mA), Fixed High (300 mA), and Adaptive (Section~\ref{sec:control_policy}). Success is defined as the gripper lifting and moving the object without visually observable slippage or damage. Failure is defined as the object fracturing or slipping, or the model aborting the process (false trigger).

\begin{figure}[b]
    \vspace{-0.4cm}
    \centering
    \includegraphics[width=\linewidth]{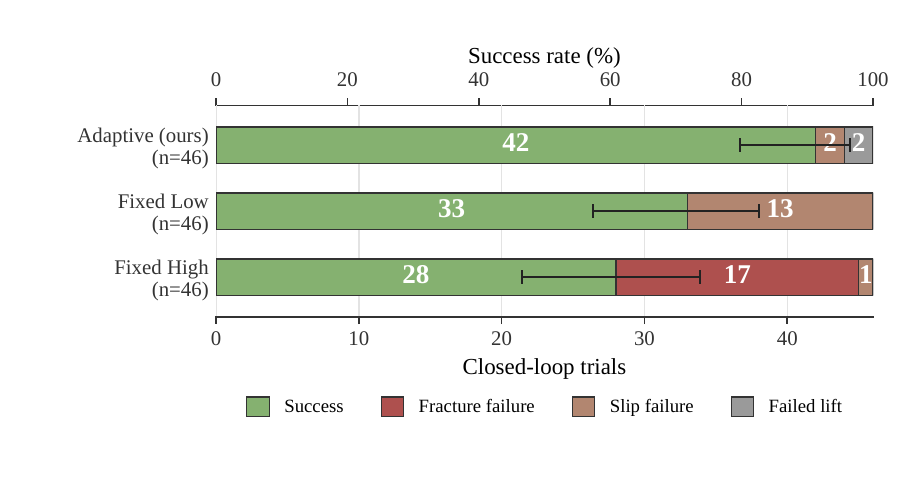}
    \vspace{-0.8cm}
    \caption{Closed-loop failure modes across three grasp-force controllers (46 trials each). Fixed Low (50~mA) grips too gently and fails primarily by slipping; Fixed High (300~mA) grips too firmly and fails primarily by fracturing; Adaptive avoids both failure modes almost entirely. Whiskers show 95\% Wilson confidence intervals on the success rate.}
    \label{fig:closed_loop}
    \vspace{-0.3cm}
\end{figure}

The upper section of Table~\ref{tab:id_ood} reports per-object and overall success rates. Fig.~\ref{fig:closed_loop} breaks down failures by mode. The two fixed-current strategies fail in essentially disjoint but mechanistically predictable ways. Fixed Low never fractures whereas Fixed High never slips. Fixed-current grasping only works well in conditions where the object property is uniform and known in advance. Our adaptive strategy, closed-loop feedback on fracture and slippage, suppresses both failure modes simultaneously, for an overall success rate of 91.3\% versus 71.7\% (Fixed Low) and 60.9\% (Fixed High) without prior knowledge of object fragility. Representative closed-loop behavior is illustrated in Fig.~\ref{fig:adaptive_sequence}.

\subsection{Out-of-Distribution Generalization}
\begin{figure}[t]
    \centering
    \includegraphics[width=\linewidth]{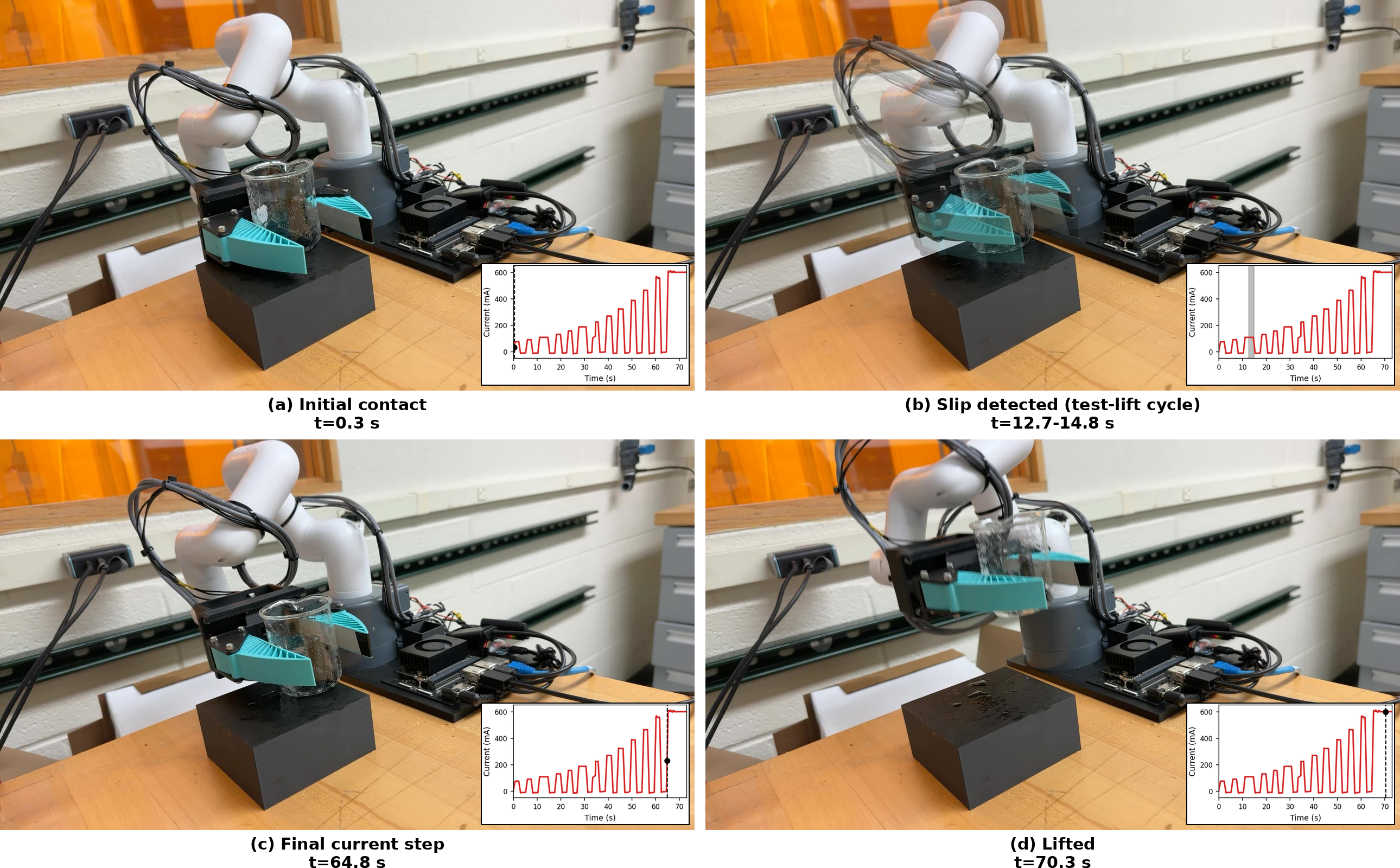}
    \vspace{-0.6cm}
    \caption{
    Representative closed-loop grasp sequence on a soap-coated glass beaker. (a) Initial contact. (b) One representative test lift (of 13 total) during which slip is detected (eight consecutive arm poses are overlaid to visualize the lifting motion while the beaker remains stationary). (c) Final current escalation after repeated slip detections. (d) Successful lift after the controller converges to a stable grasp. Insets show the commanded motor current, with the highlighted instant corresponding to each frame.
    }
    \label{fig:adaptive_sequence}
    \vspace{-0.3cm}
\end{figure}

To probe generalization beyond the training distribution, we tested the same adaptive controller on three object categories that lie outside the training data, including onion ring chips, glass beaker coated with soap (Fig.~\ref{fig:adaptive_sequence}), and plastic cup. Most notably, the glass beaker introduces both a novel material (glass, absent from all training objects) and an induced low-friction surface unlike anything the model was trained on, directly testing whether slip detection generalizes beyond specific acoustic signatures of trained materials. 

Table~\ref{tab:id_ood} (lower) reports 77.8\% (pooled across two weight conditions), 86.7\%, and 83.3\% (pooled across three weight conditions) success for the beaker, onion ring, and plastic cup categories, respectively, an aggregate 82.4\% despite no object-specific calibration or retraining. Despite the performance drop relative to in-distribution objects, the beaker still achieves 77.8\% success, suggesting that slip detection generalizes beyond material-specific acoustic patterns. Performance degrades gracefully rather than catastrophically since more lift tests were performed to reach a sufficient gripping current, and thus more inferences were made. These results suggest that the proposed sensing and control framework is beneficial for deployment in unstructured, real-world grasping scenarios.

\begin{table}[t]
    \centering
    \caption{Grasp Success Rate (\%) by Object, Weight, and Controller}
    \label{tab:id_ood}
    \vspace{-0.3cm}
    \footnotesize
    \renewcommand{\arraystretch}{1.05} 
    \setlength{\tabcolsep}{2pt} 
    \begin{tabular}{lccccc} 
        \toprule
        Object & Avg. Weight (g) & $n$ & \shortstack{Fixed Low} & \shortstack{Fixed High} & \shortstack{Adaptive} \\
        \midrule
        \multicolumn{6}{l}{\textit{In-distribution}} \\
        \midrule
        Eggshell     & 3.97 & 10 & 100.0 & 0.0   & 100.0 \\
        Chips        & 1.94 & 8  & 87.5  & 12.5  & 87.5  \\
        Water bottle & 9.6  & 4  & 100.0 & 100.0 & 100.0 \\
                     & 277  & 4  & 100.0 & 100.0 & 75.0  \\
                     & 547  & 4  & 0.0   & 100.0 & 75.0  \\
        Paper cup    & 2.3  & 4  & 100.0 & 100.0 & 100.0 \\
                     & 194  & 4  & 100.0 & 100.0 & 100.0 \\
        Soap bottle  & 178  & 4  & 0.0   & 100.0 & 100.0 \\
                     & 290  & 4  & 0.0   & 75.0  & 75.0  \\
        \midrule
        \textbf{Overall} & & 46 & \textbf{71.7} & \textbf{60.9} & \textbf{91.3} \\
        \midrule
        \midrule 
        \multicolumn{6}{l}{\textit{Out-of-distribution}} \\
        \midrule
        Onion ring     & 2.20   & 15 & & & 86.7 \\
        Beaker coated  & 50.55  & 7  & & & 100.0 \\ 
        \hspace{1em}with soap & 166.82 & 11 & & & 63.6 \\ 
        Plastic cup    & 3.83   & 5  & & & 100.0 \\
                       & 122.76 & 6  & & & 83.3  \\
                       & 241.77 & 7  & & & 71.4  \\
        \midrule
        \textbf{Overall} & & 51 & & & \textbf{82.4} \\
        \bottomrule
    \end{tabular}
    \vspace{-0.5cm}
\end{table}

\section{Conclusion}
This paper presented SAFE, a unified acoustic sensing framework for real-time slip and fracture detection in robotic grasping. By combining two low-cost passive PVDF sensors with motor proprioception on a compliant Fin Ray gripper, SAFE recognizes three grasp states (normal, slip, and fracture) using a single HistGradientBoosting classifier and integrates directly into adaptive closed-loop grasp control. The proposed framework requires neither prior knowledge of object material nor expensive sensing hardware, making it well suited for deployment in unstructured environments where object properties are unknown in advance.

Experimental results demonstrate that SAFE achieves an Alert-F1 of 0.884 with near-zero confusion between slip and fracture under Leave-One-Grasp-Out evaluation. Integrated into a real-time adaptive controller, the proposed framework achieves a 91.3\% closed-loop grasp success rate across diverse objects while maintaining 82.4\% success on previously unseen objects, all running at approximately 104 Hz on a single NVIDIA Jetson Orin Nano. These results demonstrate that reliable failure awareness for robotic grasping can be achieved using low-cost passive acoustic sensing without object-specific calibration.

Beyond the reported experiments, SAFE establishes a practical failure-aware sensing layer for robotic manipulation. It can operate as a standalone adaptive grasp controller, while the low cost, flexibility, and ease of integration of PVDF sensors also make it suitable as a plug-and-play sensing layer for existing robotic grippers and grasp-control strategies without substantial hardware redesign.

Several limitations define the current scope of this work. The training dataset spans two fracture objects and three slip objects. The out-of-distribution experiments show that the learned acoustic representation generalizes well beyond these specific categories, and extending this evaluation to a broader range of materials, including ductile or non-brittle failure modes, is a natural next step. Similarly, several per-condition trial counts in the closed-loop and out-of-distribution evaluations are modest ($n=4$--$15$). The aggregate success rates in Table~\ref{tab:id_ood} are supported by the full trial count, and expanding per-object trials would sharpen the individual comparisons further.

Looking forward, future work will focus on collecting larger and more diverse datasets, extending the framework to multi-finger PVDF configurations and different gripper types, and investigating physics-informed end-to-end learning architectures. We also plan to analyze the magnitude of slip and fracture severity for faster convergence of the grasp controller. Beyond event detection, real-time fracture anticipation is a promising direction for enabling intervention before irreversible damage occurs. More broadly, the grasp states estimated by SAFE may be tokenized for future Vision-Language-Action systems, allowing grasp outcomes to become part of higher-level manipulation planning and decision making. Together, we envision SAFE as a step toward more perceptive and failure-aware robotic manipulation systems that can not only react to grasp failures, but ultimately anticipate and reason over them.





\section*{Acknowledgment}
This work was performed during, and supported by, the Vanderbilt University School of Engineering (VUSE) Summer Research Program. The authors thank the Department of Electrical and Computer Engineering at Vanderbilt for lab and facilities support, and the Wond'ry at Vanderbilt for access to the Stratasys 3D printers. The work was supported by NSF Award No. 2334968. Claude (claude.ai, Anthropic) was used for code debugging and language polishing throughout the manuscript. The authors carefully reviewed all edits and ensured accuracy and integrity.

\bibliographystyle{IEEEtran}
\bibliography{references-final}

@misc{waltersson2026deformable,
  author       = {Waltersson, Gabriel Arslan and Karayiannidis, Yiannis},
  title        = {Deformable in-hand slip-aware tactile sensor with integrated
                  velocity, force/torque, and pressure map sensing},
  howpublished = {ar{X}iv:2606.11952},
  month        = jun,
  year         = {2026},
  note         = {Submitted to IEEE Robot. Autom. Lett.}
}

@misc{ayral2026reactive,
  author       = {Ayral, Th{\'e}o and Aloui, Saifeddine and Grossard, Mathieu},
  title        = {Reactive slip control in multifingered grasping: Hybrid
                  tactile sensing and internal-force optimization},
  howpublished = {ar{X}iv:2602.16127},
  month        = feb,
  year         = {2026},
  note         = {Accepted to IEEE Int. Conf. Robot. Autom. (ICRA)}
}

@article{shang2026forte,
  author  = {Shang, Siqi and Seo, Mingyo and Zhu, Yuke and Chin, Lillian},
  title   = {{FORTE}: Tactile force and slip sensing on compliant fingers for
             delicate manipulation},
  journal = {IEEE Robot. Autom. Lett.},
  volume  = {11},
  number  = {4},
  pages   = {4473--4480},
  month   = apr,
  year    = {2026}
}

@article{nishimura2019fluid,
  author  = {Nishimura, Toshihiro and Suzuki, Yosuke and Tsuji, Tokuo and
             Watanabe, Tetsuyou},
  title   = {Fluid pressure monitoring-based strategy for delicate grasping of
             fragile objects by a robotic hand with fluid fingertips},
  journal = {Sensors},
  volume  = {19},
  number  = {4},
  pages   = {782},
  month   = feb,
  year    = {2019}
}

@article{ishikawa2022learning,
  author  = {Ishikawa, Reina and Hamaya, Masashi and von Drigalski, Felix and
             Tanaka, Kazutoshi and Hashimoto, Atsushi},
  title   = {Learning by breaking: Food fracture anticipation for robotic food
             manipulation},
  journal = {IEEE Access},
  volume  = {10},
  pages   = {99321--99329},
  year    = {2022}
}

@inproceedings{yagawa2023learning,
  author    = {Yagawa, Rinto and Ishikawa, Reina and Hamaya, Masashi and
               Tanaka, Kazutoshi and Hashimoto, Atsushi and Saito, Hideo},
  title     = {Learning food picking without food: Fracture anticipation by
               breaking reusable fragile objects},
  booktitle = {Proc. IEEE Int. Conf. Robot. Autom. (ICRA)},
  address   = {London, U.K.},
  pages     = {917--923},
  month     = may,
  year      = {2023}
}

@incollection{chuluunbat2018acoustic,
  author    = {Chuluunbat, Turbadrakh and Kostryzhev, Andrii},
  title     = {Acoustic emission monitoring of fracture tests},
  booktitle = {Advanced Engineering Testing},
  editor    = {Ali, Aidy},
  chapter   = {2},
  publisher = {IntechOpen},
  address   = {London, U.K.},
  month     = oct,
  year      = {2018}
}

@misc{Nikafrooz2022,
  author       = {Nikafrooz, Negin and Fuge, Zachary and Leonessa, Alexander},
  title        = {Grasp control of a cable-driven robotic hand using a {PVDF}
                  slip detection sensor},
  howpublished = {ar{X}iv:2202.06140},
  month        = feb,
  year         = {2022}
}

@article{Rosle2025,
  author  = {Rosle, Muhammad Hisyam and Saffiai, Abdul Rashid and Nasir, Abdul
             and Saniman, Muhammad Nur Farhan},
  title   = {{PVDF}-based flexible piezoelectric tactile sensor for slip
             estimation using robotic gripper},
  journal = {Front. Robot. AI},
  volume  = {12},
  pages   = {1691688},
  month   = dec,
  year    = {2025}
}

@article{yuan2017gelsight,
  author  = {Yuan, Wenzhen and Dong, Siyuan and Adelson, Edward H.},
  title   = {{GelSight}: High-resolution robot tactile sensors for estimating
             geometry and force},
  journal = {Sensors},
  volume  = {17},
  number  = {12},
  pages   = {2762},
  month   = dec,
  year    = {2017}
}

@article{lambeta2020digit,
  author  = {Lambeta, Mike and Chou, Po-Wei and Tian, Stephen and Yang, Brian
             and Maloon, Benjamin and Most, Victoria Rose and Stroud, Dave and
             Santos, Raymond and Byagowi, Ahmad and Kammerer, Gregg and
             Jayaraman, Dinesh and Calandra, Roberto},
  title   = {{DIGIT}: A novel design for a low-cost compact high-resolution
             tactile sensor with application to in-hand manipulation},
  journal = {IEEE Robot. Autom. Lett.},
  volume  = {5},
  number  = {3},
  pages   = {3838--3845},
  month   = jul,
  year    = {2020}
}

@inproceedings{liu2022gelsight,
  author    = {Liu, Sandra Q. and Adelson, Edward H.},
  title     = {{GelSight Fin Ray}: Incorporating tactile sensing into a soft
               compliant robotic gripper},
  booktitle = {Proc. IEEE Int. Conf. Soft Robot. (RoboSoft)},
  address   = {Edinburgh, U.K.},
  pages     = {925--931},
  month     = apr,
  year      = {2022}
}

@article{liang2025alltact,
  author  = {Liang, Siwei and Guan, Yixuan and Xu, Jing and Qian, Hongyu and
             Zhang, Xiangjun and Wu, Dan and Ding, Wenbo and Chen, Rui},
  title   = {{AllTact Fin Ray}: A compliant robot gripper with
             omni-directional tactile sensing},
  journal = {IEEE Trans. Autom. Sci. Eng.},
  year    = {2025},
  note    = {Early access}
}

@article{nam2025tacfinray,
  author  = {Nam, Saekwang and Deng, Bowen and Lee, Loong Yi and
             Rossiter, Jonathan M. and Lepora, Nathan F.},
  title   = {{TacFinRay}: Soft tactile {Fin-Ray} finger with indirect tactile
             sensing for robust grasping},
  journal = {IEEE Robot. Autom. Lett.},
  volume  = {11},
  number  = {3},
  pages   = {2722--2729},
  month   = mar,
  year    = {2026}
}

@misc{wang2024gelsight,
  author       = {Wang, Yanzhe and Wu, Hao and Guo, Haotian and Dong, Huixu},
  title        = {{GelSight FlexiRay}: Breaking planar limits by harnessing
                  large deformations for flexible, full-coverage multimodal
                  sensing},
  howpublished = {ar{X}iv:2411.18979},
  month        = nov,
  year         = {2024}
}

@article{li2024when,
  author  = {Li, Shoujie and Wang, Zihan and Wu, Changsheng and Li, Xiang and
             Luo, Shan and Fang, Bin and Sun, Fuchun and Zhang, Xiao-Ping and
             Ding, Wenbo},
  title   = {When vision meets touch: A contemporary review for visuotactile
             sensors from the signal processing perspective},
  journal = {IEEE J. Sel. Topics Signal Process.},
  volume  = {18},
  number  = {3},
  pages   = {267--287},
  month   = apr,
  year    = {2024}
}

@article{melchiorre2023acoustic,
  author  = {Melchiorre, Jonathan and Manuello Bertetto, Amedeo and
             Rosso, Marco Martino and Marano, Giuseppe Carlo},
  title   = {Acoustic emission and artificial intelligence procedure for crack
             source localization},
  journal = {Sensors},
  volume  = {23},
  number  = {2},
  pages   = {693},
  month   = jan,
  year    = {2023}
}

@article{romano2011humaninspired,
  author  = {Romano, Joseph M. and Hsiao, Kaijen and Niemeyer, G{\"u}nter and
             Chitta, Sachin and Kuchenbecker, Katherine J.},
  title   = {Human-inspired robotic grasp control with tactile sensing},
  journal = {IEEE Trans. Robot.},
  volume  = {27},
  number  = {6},
  pages   = {1067--1079},
  month   = dec,
  year    = {2011}
}

@inproceedings{dong2018maintaining,
  author    = {Dong, Siyuan and Ma, Daolin and Donlon, Elliott and
               Rodriguez, Alberto},
  title     = {Maintaining grasps within slipping bounds by monitoring
               incipient slip},
  booktitle = {Proc. IEEE Int. Conf. Robot. Autom. (ICRA)},
  address   = {Montreal, QC, Canada},
  pages     = {3818--3824},
  month     = may,
  year      = {2019}
}

@article{james2021slip,
  author  = {James, Jasper Wollaston and Lepora, Nathan F.},
  title   = {Slip detection for grasp stabilization with a multifingered
             tactile robot hand},
  journal = {IEEE Trans. Robot.},
  volume  = {37},
  number  = {2},
  pages   = {506--519},
  month   = apr,
  year    = {2021}
}

@article{zhao2025universal,
  author  = {Zhao, Chuangri and Yu, Yang and Ye, Zeqi and Tian, Ziyang and
             Zhang, Yifan and Zeng, Ling-Li},
  title   = {Universal slip detection of robotic hand with tactile sensing},
  journal = {Front. Neurorobot.},
  volume  = {19},
  pages   = {1478758},
  month   = feb,
  year    = {2025}
}

@misc{yoo2026aslip,
  author       = {Yoo, Uksang and Mao, Yuemin and Oh, Jean and
                  Ichnowski, Jeffrey},
  title        = {{A-SLIP}: Acoustic sensing for continuous in-hand slip
                  estimation},
  howpublished = {ar{X}iv:2604.08528},
  month        = apr,
  year         = {2026}
}

@inproceedings{oh2025miccheck,
  author    = {Oh, Steven and Inui, Tai and Kuan, Magdeline and Lin, Jia-Yeu},
  title     = {{MicCheck}: Repurposing off-the-shelf pin microphones for easy
               and low-cost contact sensing},
  booktitle = {Proc. Int. Autom. Control Conf. (CACS)},
  address   = {Hsinchu, Taiwan},
  pages     = {1--6},
  month     = nov,
  year      = {2025}
}

@inproceedings{li2022see,
  author    = {Li, Hao and Zhang, Yizhi and Zhu, Junzhe and Wang, Shaoxiong
               and Lee, Michelle A. and Xu, Huazhe and Adelson, Edward and
               {Fei-Fei}, Li and Gao, Ruohan and Wu, Jiajun},
  title     = {See, hear, and feel: Smart sensory fusion for robotic
               manipulation},
  booktitle = {Proc. Conf. Robot Learn. (CoRL)},
  address   = {Auckland, New Zealand},
  month     = dec,
  year      = {2022}
}

@misc{xu2025multimodal,
  author       = {Xu, Zhuowei and Si, Zilin and Zhang, Kevin and
                  Kroemer, Oliver and Temel, Zeynep},
  title        = {A multi-modal tactile fingertip design for robotic hands to
                  enhance dexterous manipulation},
  howpublished = {ar{X}iv:2510.05382},
  month        = oct,
  year         = {2025}
}

@inproceedings{zhang2025vibecheck,
  author    = {Zhang, Kaidi and Kim, Do-Gon and Chang, Eric T. and
               Liang, Hua-Hsuan and He, Zhanpeng and Lampo, Kathryn and
               Wu, Philippe and Kymissis, Ioannis and Ciocarlie, Matei},
  title     = {{VibeCheck}: Using active acoustic tactile sensing for
               contact-rich manipulation},
  booktitle = {Proc. IEEE/RSJ Int. Conf. Intell. Robots Syst. (IROS)},
  pages     = {12278--12285},
  year      = {2025}
}

@article{Xin2016,
  author  = {Xin, Yi and Tian, Hongying and Guo, Chao and Li, Xiang and
             Sun, Hongshuai and Wang, Peiyuan and Lin, Jun and Wang, Shuhong
             and Wang, Cheng},
  title   = {{PVDF} tactile sensors for detecting contact force and slip:
             A review},
  journal = {Ferroelectrics},
  volume  = {504},
  number  = {1},
  pages   = {31--45},
  year    = {2016}
}

@article{Qi2023,
  author  = {Qi, FangXi and Xu, Lei and He, Yin and Yan, Han and Liu, Hao},
  title   = {{PVDF}-based flexible piezoelectric tactile sensors: Review},
  journal = {Cryst. Res. Technol.},
  volume  = {58},
  number  = {10},
  pages   = {2300119},
  year    = {2023}
}

@article{Zhao2026,
  author  = {Zhao, Jinwei and Dai, Shengyuan and Wang, Xin and Wang, Yongjun
             and Chen, Yinglin and Wu, Guirong and Chang, Siyuan and
             Zhao, Zhendong and Liu, Zhigui and Zhang, Haiyan and
             Xue, Chenyang and Gao, Libo},
  title   = {{PVDF}-based piezoelectric tactile sensors: Advances in material
             design, functional optimization, and applications},
  journal = {Adv. Mater. Technol.},
  volume  = {11},
  number  = {9},
  pages   = {e02194},
  year    = {2026}
}

@inproceedings{wang2026slip,
  author    = {Wang, Zerun and Kamat, Vivek and Bhansali, Shekhar},
  title     = {Investigating slip-induced frequency relationship in
               dual-channel {PVDF} sensor applied to robotic grippers},
  booktitle = {Proc. 38th Eur. Conf. Solid-State Transducers (EUROSENSORS)},
  address   = {Zurich, Switzerland},
  month     = sep,
  year      = {2026},
  note      = {To be published}
}

@inproceedings{chi2024universal,
  author    = {Chi, Cheng and Xu, Zhenjia and Pan, Chuer and Cousineau, Eric
               and Burchfiel, Benjamin and Feng, Siyuan and Tedrake, Russ and
               Song, Shuran},
  title     = {Universal manipulation interface: In-the-wild robot teaching
               without in-the-wild robots},
  booktitle = {Proc. Robot.: Sci. Syst. (RSS)},
  address   = {Delft, Netherlands},
  month     = jul,
  year      = {2024}
}

@electronic{gelsight2026digit,
  author = {{GelSight}},
  title  = {{DIGIT} tactile sensor},
  year   = {2026},
  url    = {https://www.gelsight.com/product/digit-tactile-sensor/},
  note   = {Accessed: Jun. 3, 2026}
}

@electronic{gelsight2026mini,
  author = {{GelSight}},
  title  = {{GelSight Mini} system},
  year   = {2026},
  url    = {https://www.gelsight.com/product/gelsight-mini-system/},
  note   = {Accessed: Jun. 3, 2026}
}

\end{document}